\documentclass[11pt,a4paper]{article}
\usepackage[hyperref]{naaclhlt2018}
\usepackage{times}
\usepackage{latexsym}
\usepackage{graphicx}
\usepackage{url}

\aclfinalcopy

\title{UTFPR at SemEval-2019 Task 5: \\ Hate Speech Identification with Recurrent Neural Networks}

\author{
  Gustavo Henrique Paetzold\textsuperscript{1}, Shervin Malmasi\textsuperscript{2},  Marcos Zampieri\textsuperscript{3} \\
  \textsuperscript{1}Universidade Tecnol\'{o}gica Federal do Paran\'{a}, Toledo-PR, Brazil\\
  \textsuperscript{2}Harvard Medical School, Boston, United States\\
  \textsuperscript{3}University of Wolverhampton, Wolverhampton, United Kingdom\\
  {\tt ghpaetzold@utfpr.edu.br} 
}

\date{}

\begin{document}
\maketitle
\begin{abstract}
  In this paper we revisit the problem of automatically identifying hate speech in posts from social media. We approach the task using a system based on minimalistic compositional Recurrent Neural Networks (RNN). We tested our approach on the SemEval-2019 Task 5: Multilingual Detection of Hate Speech Against Immigrants and Women in Twitter (HatEval) shared task dataset. The dataset made available by the HatEval organizers contained English and Spanish posts retrieved from Twitter annotated with respect to the presence of hateful content and its target. In this paper we present the results obtained by our system in comparison to the other entries in the shared task. Our system achieved competitive performance ranking 7\textsuperscript{th} in sub-task A out of 62 systems in the English track.
\end{abstract}

\section{Introduction}

Abusive and offensive content such as aggression, cyberbulling, and hate speech have become pervasive in social media. The widespread of offensive content in social media is a reason of concern for governments worldwide and technology companies, which have been heavily investing in ways to cope with such content using human moderation of posts, triage of content, deletion of offensive posts, and banning abusive users. 

One of the most common and effective strategies to tackle the problem of offensive language online is to train systems capable of recognizing such content. Several studies have been published in the last few years on identifying abusive language \cite{nobata2016abusive}, cyber aggression \cite{trac2018report}, cyber bullying \cite{dadvar2013improving}, and hate speech \cite{burnap2015cyber,davidson2017automated}. As evidenced in two recent surveys \cite{schmidt2017survey,fortuna2018survey} and in a number of other studies \cite{malmasi2017detecting,gamback2017using,elsherief2018hate,zhang2018detecting}, the identification of hate speech is the most popular of what \newcite{waseem2017understanding} refers to as ``abusive language detection sub-tasks''. 

This paper deals with the hate speech identification in English and Spanish posts from social media. We present our submissions to the SemEval-2019 Task 5: Multilingual Detection of Hate Speech Against Immigrants and Women in Twitter (HatEval) shared task. We participated in sub-task A which is a binary classification task in which systems are trained to discriminate between posts containing hate speech and posts which do not contain any form of hate speech. Our approach, presented in detail in Section \ref{sec:approach}, combines compositional Recurrent Neural Networks (RNN) and transfer learning and achieved competitive performance in the shared task.

\section{Related Work}

As evidenced in the introduction of this paper, there have been a number of studies on automatic hate speech identification published in the last few years. One of the most influential recent papers on hate speech identification is the one by \newcite{davidson2017automated}. In this paper, the authors presented the Hate Speech Detection dataset which contains posts retrieved from social media labeled with three categories: OK (posts not containing profanity or hate speech), Offensive (posts containing swear words and general profanity), and Hate (posts containing hate speech). It has been noted in \newcite{davidson2017automated}, and in other works \cite{malmasi2018challenges}, that training models to discriminate between general profanity and hate speech is far from trivial due to, for example, the fact that a significant percentage of hate speech posts contain swear words. It has been argued that annotating texts with respect to the presence of hate speech has an intrinsic degree of subjectivity \cite{malmasi2018challenges}.

Along with the recent studies published, there have been a few related shared tasks organized on the topic. These include GermEval \cite{wiegand2018overview} for German, TRAC \cite{trac2018report} for English and Hindi, and OffensEval\footnote{\url{https://competitions.codalab.org/competitions/20011}} \cite{offenseval} for English. The latter is also organized within the scope of SemEval-2019. OffensEval considers offensive language in general whereas HatEval focuses on hate speech.

\newcite{waseem2017understanding} proposes a typology of abusive language detection sub-tasks taking two factors into account: the target of the message and whether the content is explicit or implicit. Considering that hate speech is commonly understood as speech attacking a group based on ethnicity, religion, etc, and that cyber bulling, for example, is understood as an attack towards an individual, the target factor plays an important role in the identification and the definition of hate speech when compared to other forms of abusive content. 

The two SemEval-2019 shared tasks, HatEval and OffensEval, both include a sub-task on target identification as discussed in \newcite{waseem2017understanding}. HatEval includes the target annotation in its sub-task B with two classes (individual or group) whereas OffensEval includes it in its sub-task C with three classes (individual, group or others). Another important similarity between these two tasks is that both include a more basic binary classification task in sub-task A. In HatEval, posts are labeled as as to whether they contain hate speech or not and in OffensEval, posts are labeled as being offensive or not. As OffensEval considers multiple types of offensive contents, the hierarchical annotation model used to annotate OLID \cite{OLID}, the dataset used in OffensEval, includes an annotation level distinguishing between the type of offensive content that posts include with two classes: insults and threats, and general profanity. This type annotation is used in OffensEval's sub-task B.

\section{Task Description}

%The presented contributions are systems submitted to the HatEval shared task of SemEval-2019. 
HatEval \cite{hateval2019semeval} provides participants with annotated datasets to create systems capable of properly identifying hate speech in tweets written in both English and Spanish. 

The training, development, trial, and test sets provided for English are composed of 9,000, 1,000, 100 and 3,000 instances, respectively. The training, development, trial and test sets provided for Spanish are composed of 4,500, 500, 100 and 1,600 instances, respectively. Each instance is composed of a tweet and three binary labels: One that indicates whether or not hate speech is featured in the tweet, one indicating whether the hate speech targets a group or an individual, and another indicating whether or not the author of the tweet is aggressive. HatEval has 2 sub-tasks:

\begin{itemize}
	\item \textbf{Sub-task A:} Judging whether or not a tweet is hateful.
	\item \textbf{Sub-task B:} Correctly predicting all three of the aforementioned labels.
\end{itemize}

\noindent In this paper, we focus on Task A exclusively, for both English and Spanish. We participated in the competition using the team name UTFPR.

\section{The UTFPR Models}
\label{sec:approach}

\begin{figure*}[htbp!]
\centering    
\includegraphics[width=\textwidth]{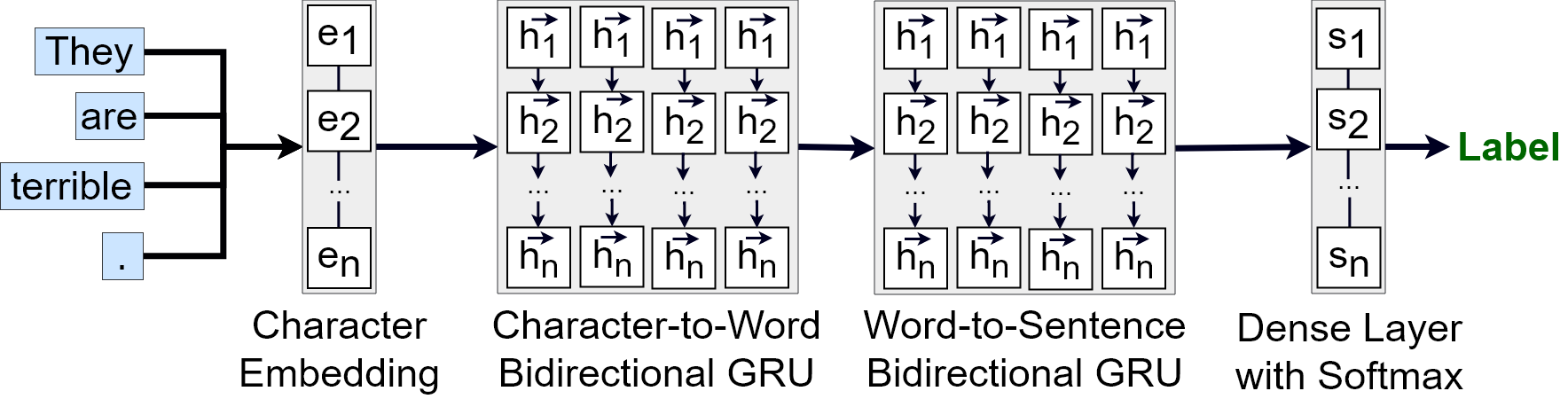}
\caption{Architecture of the UTFPR models.}
\label{fig:arch}
\end{figure*}

The UTFPR models are minimalistic Recurrent Neural Networks (RNNs) that learn compositional numerical representations of words based on the sequence of characters that compose them, then use them to learn a final representation for the sentence being analyzed. These models, of which the architecture is illustrated in Figure \ref{fig:arch}, are somewhat similar to those of \newcite{ling2015char} and \newcite{paetzold2018wassa}, who use RNNs to create compositional neural models for different tasks.

As illustrated, the UTFPR models take as input a sentence, split it into words, then split the words into a sequence of characters in order to pass them through a character embedding layer. The character embeddings are passed onto a set of bidirectional RNN layers that produces word representations, then a second set of layers produces a final representation of the sentence. Finally, this representation is passed through a softmax dense layer that produces a final classification label.

For each language, we created two variants of UTFPR: one trained exclusively over the training data provided by the organizers (UTFPR/O), and another that uses a pre-trained set of character-to-word RNN layers extracted from the models introduced by \newcite{paetzold2018wassa} (UTFPR/W). The pre-trained model was trained for the English multi-class classification Emotion Analysis shared task of WASSA 2018, which featured a training set of $153,383$ instances composed of a tweet and an emotion label. This pre-trained model for English was used for the UTFPR/W variant of both languages, since we wanted to test the hypothesis that pre-training a character-to-word RNN on a large dataset for English can improve the performance of compositional models for both English and Spanish.

We use 25 dimensions for the size of our character embeddings, and two layers of Gated Recurrent Units for our bidirectional RNNs with 60 hidden nodes each and 50\% dropout. We saved a model after each training iteration and picked the one with the lowest error on the development set. The UTFPR/W model went through the same training process as UTFPR/O, with the pre-trained character-to-word RNN layers being fine-tuned for the task at hand.

Table \ref{table:trial} showcases the F-scores obtained by the UTFPR systems on the trial set of Task A. Because of its superior performance, we chose to submit the UTFPR/W variants as our official entry.

\begin{table}[htpb]
%\setlength{\tabcolsep}{2.8pt}
%\small
\centering
\begin{tabular}{c|cc}
 & \multicolumn{2}{|c}{\textbf{F-scores}} \\
\textbf{System} & \textbf{English} & \textbf{Spanish} \\
\hline
UTFPR/O & $0.509$ & $0.601$ \\
UTFPR/W & $0.570$ & $0.665$ \\
\end{tabular}
\caption{F-scores obtained for the trial set at HatEval Task A for both languages.}
\label{table:trial}
\end{table}

\section{Results and Discussion}

\subsection{Shared Task Performance}

Tables \ref{table:testen} and \ref{table:testes} feature the F-scores obtained by the UTFPR systems and the 3 best and worst performing systems at HatEval Task A for English and Spanish, respectively. Ultimately, the UTFPR/W systems submitted ranked 7\textsuperscript{th} out of 62 valid submissions for English, and 31\textsuperscript{st} out of 35 valid submissions for Spanish.

\begin{table}[ht]
%\setlength{\tabcolsep}{2.8pt}
%\small
\centering
\begin{tabular}{c|c}
\textbf{System} & \textbf{F-scores} \\
\hline
FERMI & $0.650$ \\
Panaetius & $0.570$ \\
YNU\_DYX & $0.550$ \\
\hline
UTFPR/O & $0.524$ \\
UTFPR/W & $0.513$ \\
\hline
MELODI & $0.350$ \\
INGEOTEC & $0.350$ \\
INAOE-CIMAT & $0.340$ \\
\end{tabular}
\caption{F-scores obtained at HatEval Task A for the English language. At the top and bottom of the table are featured the top and bottom 3 systems submitted to the shared task, respectively.}
\label{table:testen}
\end{table}

\begin{table}[htpb]
%\setlength{\tabcolsep}{2.8pt}
%\small
\centering
\begin{tabular}{c|c}
\textbf{System} & \textbf{F-scores} \\
\hline
mineriaUNAM & $0.730$ \\
Atalaya & $0.730$ \\
MITRE & $0.730$ \\
\hline
UTFPR/O & $0.664$ \\
UTFPR/W & $0.636$ \\
\hline
jhouston & $.630$ \\
LU team & $0.620$ \\
TuEval & $0.620$ \\
\end{tabular}
\caption{F-scores obtained at HatEval Task A for the Spanish language. At the top and bottom of the table are featured the top and bottom 3 systems submitted to the shared task, respectively.}
\label{table:testes}
\end{table}

\begin{figure*}[htbp!]
\centering    
\includegraphics[width=\textwidth]{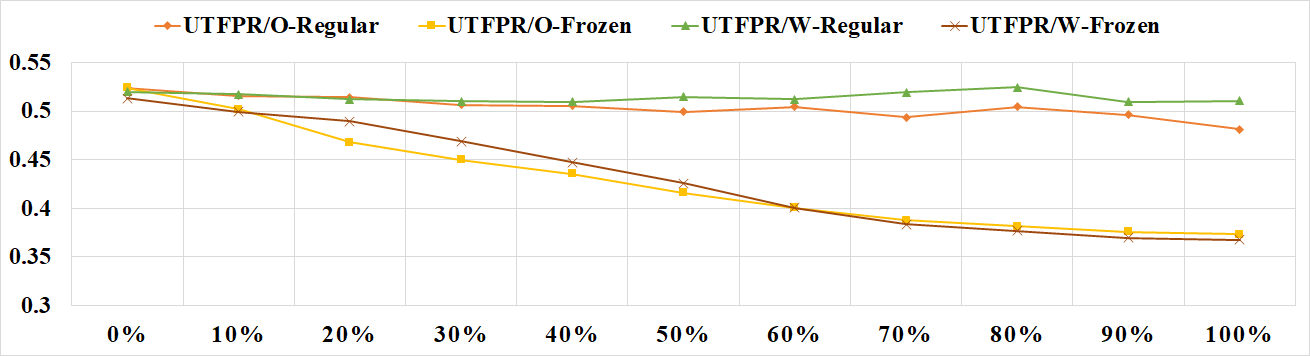}
\caption{F-scores of our robustness experiments for English. The horizontal axis represents the proportion of noisy words in the input sentence, and the vertical axis the F-scores.}
\label{fig:jammeden}
\end{figure*}

\begin{figure*}[htbp!]
\centering    
\includegraphics[width=\textwidth]{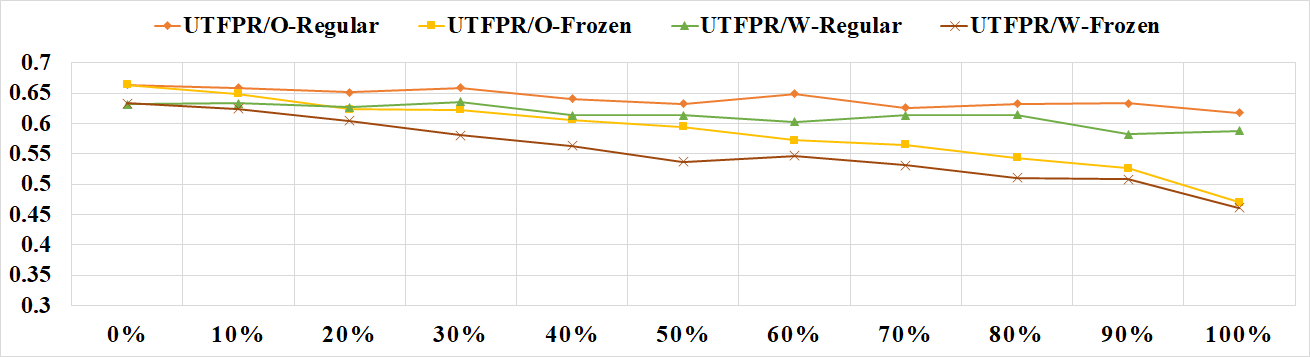}
\caption{F-scores of our robustness experiments for Spanish. The horizontal axis represents the proportion of noisy words in the input sentence, and the vertical axis the F-scores.}
\label{fig:jammedes}
\end{figure*}

\noindent One of the aspects we wanted to test with our participation in this shared task was the extent to which pre-training a character-to-word RNN over a larger dataset for an analogous task helped the models. Our results show that, even though using a pre-trained RNN considerably improved the performance of our models in the trial experiments, it actually compromised their performance for the test set a little. We believe that this was caused because the development set was more representative of the trial than the test set. Overall, submitting UTFPR/W instead of UTFPR/O cost us 2 ranks for English and 3 for Spanish.

\subsection{Robustness Assessment}
\label{robustness}

In order to test the robustness of the UTFPR systems, we had to generate different noisy versions of the test set with increasing volumes of noise artificially added to them.

To do so, we introduced a modification to $N$\% of randomly selected words in each sentence in the datasets. The modifications could be either the deletion of a randomly selected character  ($50$\% chance) or its duplication ($50$\% chance). We used $0\!\leq \! N\!\leq \! 100$ in intervals of 10, resulting in a total of 11 increasingly noisy versions. The next step was to create ``frozen'' versions of the UTFPR models that act as if any word out of the training set's vocabulary is unknown. If a word of the test set is not present in the vocabulary of the training set, it produces a numerical vector full of 1's that represents an out-of-vocabulary word.

Figures \ref{fig:jammeden} and \ref{fig:jammedes} show the results obtained for English and Spanish, respectively. As it can be noticed, our compositional models are much more robust than the frozen alternatives, suffering very faint losses in F-score even when 100\% of the words in the input sentence are noisy.

\section{Conclusions}

In this contribution, we presented the UTFPR systems submitted to the HatEval 2019 shared task. The systems are based on compositional RNN models trained exclusively over the training data provided by the organizers. We introduced two variants of our models: one trained entirely on the shared task's data (UTFPR/O), and another with a set of pre-trained character-to-word RNN layers fine-tuned to the task at hand (UTFPR/W). Our results show that, despite its simplicity, the UTFPR/O model attained competitive results for English, placing it 7\textsuperscript{th} out of 62 submissions. Furthermore, the results of this shared task indicate that our models are very robust, being able to handle even substantially noisy inputs. In the future, we intend to test more reliable ways of re-using pre-trained compositional models.

\section*{Acknowledgements}

We would like to thank the organizers of the HatEval shared task for providing participants with this dataset and for organizing this interesting shared task. We gratefully acknowledge the support of NVIDIA Corporation with the donation of the Titan V GPU used for this research.

\bibliography{references}

\begin{thebibliography}{19}
\expandafter\ifx\csname natexlab\endcsname\relax\def\natexlab#1{#1}\fi

\bibitem[{Basile et~al.(2019)Basile, Bosco, Fersini, Nozza, Patti, Rangel,
  Rosso, and Sanguinetti}]{hateval2019semeval}
Valerio Basile, Cristina Bosco, Elisabetta Fersini, Debora Nozza, Viviana
  Patti, Francisco Rangel, Paolo Rosso, and Manuela Sanguinetti. 2019.
\newblock Semeval-2019 task 5: Multilingual detection of hate speech against
  immigrants and women in twitter.
\newblock In \emph{Proceedings of the 13th International Workshop on Semantic
  Evaluation (SemEval-2019)}. Association for Computational Linguistics.

\bibitem[{Burnap and Williams(2015)}]{burnap2015cyber}
Pete Burnap and Matthew~L Williams. 2015.
\newblock Cyber hate speech on twitter: An application of machine
  classification and statistical modeling for policy and decision making.
\newblock \emph{Policy \& Internet}, 7(2):223--242.

\bibitem[{Dadvar et~al.(2013)Dadvar, Trieschnigg, Ordelman, and
  de~Jong}]{dadvar2013improving}
Maral Dadvar, Dolf Trieschnigg, Roeland Ordelman, and Franciska de~Jong. 2013.
\newblock Improving cyberbullying detection with user context.
\newblock In \emph{Advances in Information Retrieval}, pages 693--696.
  Springer.

\bibitem[{Davidson et~al.(2017)Davidson, Warmsley, Macy, and
  Weber}]{davidson2017automated}
Thomas Davidson, Dana Warmsley, Michael Macy, and Ingmar Weber. 2017.
\newblock {Automated Hate Speech Detection and the Problem of Offensive
  Language}.
\newblock In \emph{Proceedings of ICWSM}.

\bibitem[{ElSherief et~al.(2018)ElSherief, Kulkarni, Nguyen, Wang, and
  Belding}]{elsherief2018hate}
Mai ElSherief, Vivek Kulkarni, Dana Nguyen, William~Yang Wang, and Elizabeth
  Belding. 2018.
\newblock {Hate Lingo: A Target-based Linguistic Analysis of Hate Speech in
  Social Media}.
\newblock \emph{arXiv preprint arXiv:1804.04257}.

\bibitem[{Fortuna and Nunes(2018)}]{fortuna2018survey}
Paula Fortuna and S{\'e}rgio Nunes. 2018.
\newblock {A Survey on Automatic Detection of Hate Speech in Text}.
\newblock \emph{ACM Computing Surveys (CSUR)}, 51(4):85.

\bibitem[{Gamb{\"a}ck and Sikdar(2017)}]{gamback2017using}
Bj{\"o}rn Gamb{\"a}ck and Utpal~Kumar Sikdar. 2017.
\newblock {Using Convolutional Neural Networks to Classify Hate-speech}.
\newblock In \emph{Proceedings of the First Workshop on Abusive Language
  Online}, pages 85--90.

\bibitem[{Kumar et~al.(2018)Kumar, Ojha, Malmasi, and
  Zampieri}]{trac2018report}
Ritesh Kumar, Atul~Kr. Ojha, Shervin Malmasi, and Marcos Zampieri. 2018.
\newblock {Benchmarking Aggression Identification in Social Media}.
\newblock In \emph{Proceedings of the First Workshop on Trolling, Aggression
  and Cyberbulling (TRAC)}, Santa Fe, USA.

\bibitem[{Ling et~al.(2015)Ling, Dyer, Black, Trancoso, Fermandez, Amir,
  Marujo, and Luis}]{ling2015char}
Wang Ling, Chris Dyer, Alan~W Black, Isabel Trancoso, Ramon Fermandez, Silvio
  Amir, Luis Marujo, and Tiago Luis. 2015.
\newblock Finding function in form: Compositional character models for open
  vocabulary word representation.
\newblock In \emph{Proceedings of the 2015 EMNLP}, pages 1520--1530.
  Association for Computational Linguistics.

\bibitem[{Malmasi and Zampieri(2017)}]{malmasi2017detecting}
Shervin Malmasi and Marcos Zampieri. 2017.
\newblock {Detecting Hate Speech in Social Media}.
\newblock In \emph{Proceedings of the International Conference Recent Advances
  in Natural Language Processing (RANLP)}, pages 467--472.

\bibitem[{Malmasi and Zampieri(2018)}]{malmasi2018challenges}
Shervin Malmasi and Marcos Zampieri. 2018.
\newblock {Challenges in Discriminating Profanity from Hate Speech}.
\newblock \emph{Journal of Experimental \& Theoretical Artificial
  Intelligence}, 30:1--16.

\bibitem[{Nobata et~al.(2016)Nobata, Tetreault, Thomas, Mehdad, and
  Chang}]{nobata2016abusive}
Chikashi Nobata, Joel Tetreault, Achint Thomas, Yashar Mehdad, and Yi~Chang.
  2016.
\newblock {Abusive Language Detection in Online User Content}.
\newblock In \emph{Proceedings of the 25th International Conference on World
  Wide Web}, pages 145--153. International World Wide Web Conferences Steering
  Committee.

\bibitem[{Paetzold(2018)}]{paetzold2018wassa}
Gustavo Paetzold. 2018.
\newblock Utfpr at iest 2018: Exploring character-to-word composition for
  emotion analysis.
\newblock In \emph{Proceedings of the 9th EMNLP}, pages 176--181. Association
  for Computational Linguistics.

\bibitem[{Schmidt and Wiegand(2017)}]{schmidt2017survey}
Anna Schmidt and Michael Wiegand. 2017.
\newblock {A Survey on Hate Speech Detection Using Natural Language
  Processing}.
\newblock In \emph{Proceedings of the Fifth International Workshop on Natural
  Language Processing for Social Media. Association for Computational
  Linguistics}, pages 1--10, Valencia, Spain.

\bibitem[{Waseem et~al.(2017)Waseem, Davidson, Warmsley, and
  Weber}]{waseem2017understanding}
Zeerak Waseem, Thomas Davidson, Dana Warmsley, and Ingmar Weber. 2017.
\newblock {Understanding Abuse: A Typology of Abusive Language Detection
  Subtasks}.
\newblock In \emph{Proceedings of the First Workshop on Abusive Langauge
  Online}.

\bibitem[{Wiegand et~al.(2018)Wiegand, Siegel, and
  Ruppenhofer}]{wiegand2018overview}
Michael Wiegand, Melanie Siegel, and Josef Ruppenhofer. 2018.
\newblock {Overview of the GermEval 2018 Shared Task on the Identification of
  Offensive Language}.
\newblock In \emph{Proceedings of GermEval}.

\bibitem[{Zampieri et~al.(2019{\natexlab{a}})Zampieri, Malmasi, Nakov,
  Rosenthal, Farra, and Kumar}]{OLID}
Marcos Zampieri, Shervin Malmasi, Preslav Nakov, Sara Rosenthal, Noura Farra,
  and Ritesh Kumar. 2019{\natexlab{a}}.
\newblock {Predicting the Type and Target of Offensive Posts in Social Media}.
\newblock In \emph{Proceedings of NAACL}.

\bibitem[{Zampieri et~al.(2019{\natexlab{b}})Zampieri, Malmasi, Nakov,
  Rosenthal, Farra, and Kumar}]{offenseval}
Marcos Zampieri, Shervin Malmasi, Preslav Nakov, Sara Rosenthal, Noura Farra,
  and Ritesh Kumar. 2019{\natexlab{b}}.
\newblock {SemEval-2019 Task 6: Identifying and Categorizing Offensive Language
  in Social Media (OffensEval)}.
\newblock In \emph{Proceedings of The 13th International Workshop on Semantic
  Evaluation (SemEval)}.

\bibitem[{Zhang et~al.(2018)Zhang, Robinson, and Tepper}]{zhang2018detecting}
Ziqi Zhang, David Robinson, and Jonathan Tepper. 2018.
\newblock {Detecting Hate Speech on Twitter Using a Convolution-GRU Based Deep
  Neural Network}.
\newblock In \emph{Lecture Notes in Computer Science}. Springer Verlag.

\end{thebibliography}
\bibliographystyle{acl_natbib}

\end{document}